\documentclass[11pt,a4paper]{article}

\usepackage{preprint}
\usepackage[authoryear,round]{natbib}

\usepackage{amsmath,amsfonts,bm}

\def\eqref#1{equation~\ref{#1}}

\def\1{\bm{1}}

\DeclareMathAlphabet{\mathsfit}{\encodingdefault}{\sfdefault}{m}{sl}
\SetMathAlphabet{\mathsfit}{bold}{\encodingdefault}{\sfdefault}{bx}{n}

\usepackage{graphicx}
\usepackage{booktabs}
\usepackage{array}
\usepackage{xltabular}
\usepackage{multirow}
\usepackage{placeins}
\usepackage{amssymb}
\usepackage{xcolor}
\usepackage{hyperref}
\usepackage{url}
\usepackage{fvextra}
\usepackage{enumitem}
\DeclareUnicodeCharacter{2082}{$_2$}
\DeclareUnicodeCharacter{00D7}{$\times$}
\DeclareUnicodeCharacter{2013}{--}
\DeclareUnicodeCharacter{2014}{---}
\DeclareUnicodeCharacter{2019}{'}
\DeclareUnicodeCharacter{00B0}{$^\circ$}
\DeclareUnicodeCharacter{00B1}{$\pm$}
\DeclareUnicodeCharacter{201C}{``}
\DeclareUnicodeCharacter{201D}{''}
\DeclareUnicodeCharacter{2212}{-}

\graphicspath{{figures/}}

\newcommand{\paperlogos}{%
  \makebox[\textwidth]{%
    \raisebox{-0.5\height}{\includegraphics[height=0.22in]{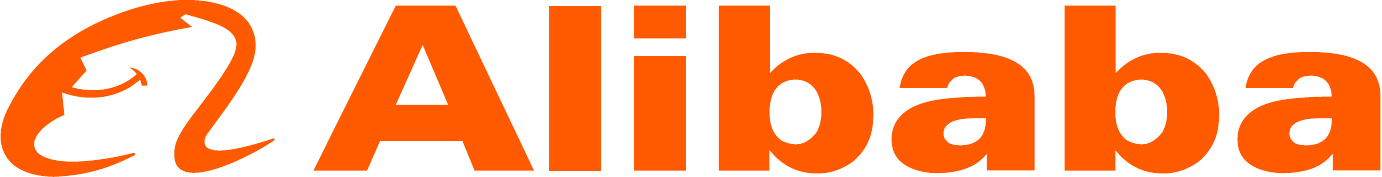}}%
    \hfill
    \raisebox{-0.5\height}{\includegraphics[height=0.34in]{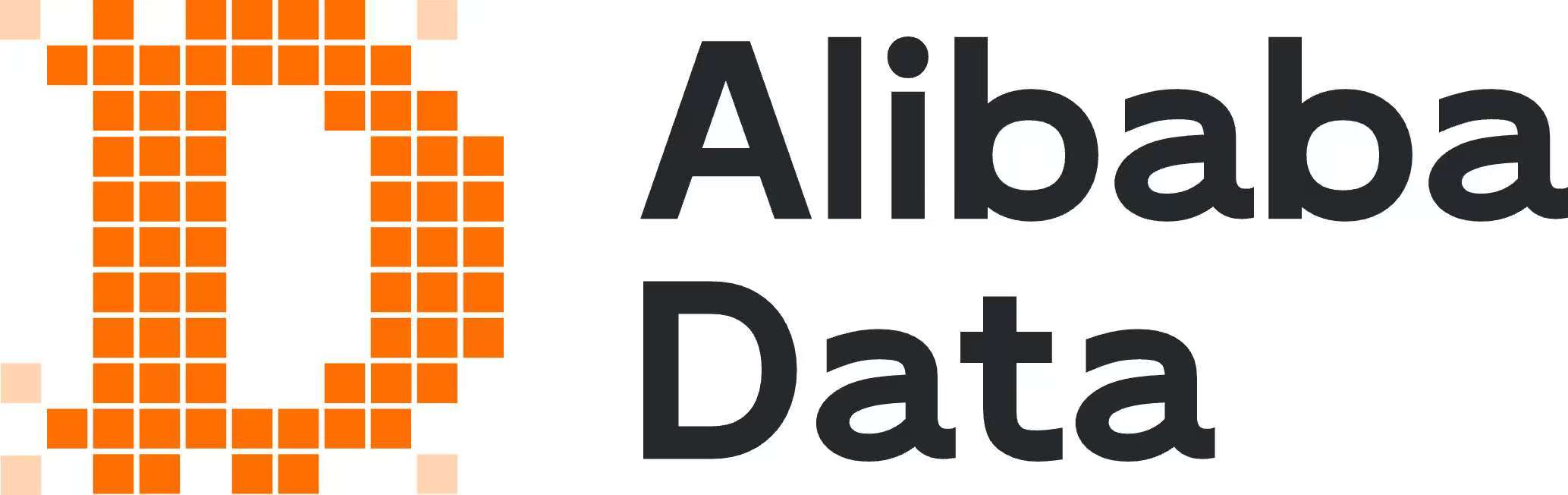}}%
  }%
}

\title{SciWorkBench: A Benchmark for\\Real Scientific Workflows}

\author{%
{\fontsize{10}{15}\selectfont
\paperauthor{Yuchen Zhang}{1,2,*}\quad \paperauthor{Taolin Han}{1,3,*}\quad \paperauthor{Jinghang Wang}{1,*}\quad \paperauthor{Jinxin Wang}{1,2}\quad \paperauthor{Yun Wu}{1}\\[2pt]\paperauthor{Wai Yuet Chiu}{1}\quad \paperauthor{Xiaogang Li}{1}\quad \paperauthor{Yuting Fu}{1}\quad \paperauthor{Kewei Sun}{1}\quad \paperauthor{Jiaxin Li}{1}\quad \paperauthor{Qile Jin}{1}\\[2pt]\paperauthor{Jiajia Li}{1}\quad \paperauthor{Ruodan Chen}{1}\quad \paperauthor{Qingteng Chen}{1}\quad \paperauthor{Jialu Nie}{1}\quad \paperauthor{Junchao Li}{1}\quad \paperauthor{Yifei Zhang}{1,4}\\[2pt]\paperauthor{Chen Zhao}{1,2}\quad \paperauthor{Hu Wei}{1}\quad \paperauthor{Lin Qu}{1}\quad \paperauthor{Yanni Yin}{5,\dagger}\quad \paperauthor{Bing Zhao}{1,\dagger}}\\[10pt]
{\fontsize{9}{12}\selectfont \authoraffiliations}\\[8pt]
{\footnotesize $^{*}$Equal contribution.\quad $^{\dagger}$Corresponding authors.}\\[4pt]
{\footnotesize \href{mailto:wjh2018@mail.ustc.edu.cn}{\nolinkurl{wjh2018@mail.ustc.edu.cn}}}
}
\date{}
\newcommand{\paperauthor}[2]{\mbox{\textbf{#1}$^{#2}$}}

\newcommand{\authoraffiliations}{%
  \mbox{$^{1}$Alibaba Group}\quad
  \mbox{$^{2}$Tsinghua University}\quad
  \mbox{$^{3}$University of Chinese Academy of Sciences}\\[3pt]
  \mbox{$^{4}$University of Alberta}\quad
  \mbox{$^{5}$Zhejiang University}%
}

\hypersetup{
  colorlinks=true,
  urlcolor=seedblue, linkcolor=seedblue, citecolor=seedblue,
  pdftitle={SciWorkBench: A Benchmark for Real Scientific Workflows},
  pdfauthor={Yuchen Zhang, Taolin Han, Jinghang Wang, Jinxin Wang, Yun Wu,
    Wai Yuet Chiu, Xiaogang Li, Yuting Fu, Kewei Sun, Jiaxin Li, Qile Jin, Jiajia Li,
    Ruodan Chen, Qingteng Chen, Jialu Nie, Junchao Li, Yifei Zhang, Chen Zhao,
    Hu Wei, Lin Qu, Yanni Yin, Bing Zhao}
}

\begin{document}

\maketitle

\begin{abstract}
Language models are increasingly being used as agents capable of planning and executing projects, as well as refining their actions through feedback. This shift raises a question: can current agents complete projects like real scientists? To examine this question, we introduce SciWorkBench, a benchmark for evaluating agents' abilities on real science tasks across biology, chemistry, materials science, and physics. SciWorkBench is built from peer-reviewed literature or expert-contributed research scenarios. We also propose a unified scientific analysis framework validated by domain experts, which can categorize most tasks in SciWorkBench. Each task covers one or more stages of this framework. We evaluate up-to-date models within the Claude Code and Codex agent frameworks under a unified execution and grading protocol. The best-performing agent achieves a mean score of 57.38, indicating that current agents still struggle with realistic scientific workflows. Our failure analysis identifies three recurring weaknesses at the workflow level: difficulty translating scientific knowledge into task-specific procedures, weak integration of evidence across different modalities, and limited capacity for scientific self-correction. These findings highlight the importance of workflow-based evaluation for measuring progress toward reliable AI systems for scientific research.
\end{abstract}

\section{Introduction}
\label{sec:introduction}

Large language models are rapidly evolving from static question-answering systems into agents that can plan, invoke tools, execute code, inspect intermediate results, and revise their actions in closed-loop environments~\citep{yao2023react,schick2023toolformer,qin2024toolllm}. This transition is important for scientific discovery, which often involves data exploration followed by hypothesis testing and refinement. Recent scientific agents have begun to connect language models with research tools, literature search, and experimental workflows~\citep{bran2024chemcrow,boiko2023coscientist,lu2024aiscientist}.

This progress raises a central evaluation question: can current agents help with scientific discovery? Existing scientific benchmarks evaluate only isolated parts of the scientific process. Static benchmarks mainly test domain knowledge and reasoning~\citep{laurent2024labbench}; data-analysis benchmarks test whether models can generate executable programs or reproduce computational analyses~\citep{tian2024scicode,chen2025scienceagentbench,siegel2024corebench}; biomedical and genomics benchmarks focus on domain-specific analysis tasks~\citep{mitchener2025bixbench,li2026genebenchpro}; and tool-use benchmarks evaluate agents in predefined scientific tool environments~\citep{bragg2026astabench,shen2026sciagentgym}. These benchmarks usually test a narrow slice of scientific capability rather than the full decision process needed to turn experimental evidence into a defensible conclusion.

The missing piece is workflow-level scientific judgment. Real laboratory analysis is not a single question-answering or tool-use episode. It is a hypothesis-driven, data-centered process in which scientists inspect raw experimental data, interpret evidence, and revise conclusions. At each step, they must decide whether the data are reliable, whether artifacts or uncertainty limit the analysis, and what conclusions are actually supported by the evidence. Throughout this process, an agent must not only execute computations, but also reason about data quality, experimental artifacts, and the evidential boundary between supported conclusions and speculative mechanisms. These workflow-level decisions are central to scientific practice but remain insufficiently captured by current benchmarks.

We introduce SciWorkBench, a benchmark for evaluating scientific agents on realistic experimental-science workflows across classical domains: biology, chemistry, materials science, and physics. Each task is built around a realistic research scenario and provides the agent with information similar to what a scientist would encounter. Central to our benchmark is a unified experimental-science workflow framework that abstracts common structures across laboratory research while preserving discipline-specific requirements.

Our contributions are summarized as follows:
\begin{itemize}
    \item \textbf{Unified scientific analysis framework.} We propose a scientific analysis framework validated by domain experts, including scientific hypothesis, raw experimental data, data curation and preprocessing, preliminary analysis, in-depth analysis, modeling and simulation or modeling and prediction, scientific conclusion, and iterative revision.
    \item \textbf{Multi-domain science benchmark.} We construct SciWorkBench as a high-quality benchmark spanning biology, chemistry, materials science, and physics, covering representative subfields and experimental research practice.
    \item \textbf{Systematic agent evaluation.} We evaluate 13 frontier model--harness agents on 98 SciWorkBench tasks under a unified execution and rubric-based grading protocol. Agents performance depends jointly on the model, harness, and scientific domains.
    \item \textbf{Analysis of scientific-agent failures.} We analyze failed cases and identify three recurring workflow-level failure patterns: procedural mismatch, weak cross-modal evidence integration, and limited scientific self-correction.
    
\end{itemize}

\section{Related Work}
\label{sec:related-work}

\paragraph{Agentic benchmarks in general domains.}
A broad line of work evaluates language models as agents in interactive digital environments, including software engineering~\citep{jimenez2024swebench}, web and operating-system control~\citep{zhou2024webarena,xie2024osworld}, heterogeneous tool-use environments~\citep{liu2024agentbench,mialon2023gaia}, and command-line task completion~\citep{tbench2026}. These benchmarks have established rigorous execution-based harnesses for agent evaluation, but their tasks are primarily drawn from software, web, and everyday digital settings rather than from scientific practice.

\paragraph{Scientific benchmarks for agents.}
Closer to our work, recent scientific agent benchmarks evaluate models on scientific coding, data-driven discovery, computational reproducibility, and domain-specific analysis tasks~\citep{chen2025scienceagentbench,siegel2024corebench,majumder2024discoverybench,tian2024scicode,mitchener2025bixbench}. Other benchmarks focus on agents for machine-learning research and paper replication~\citep{chan2024mlebench,huang2024mlagentbench,starace2025paperbench}. More recent efforts such as Terminal-Bench-Science~\citep{tbenchscience2026} and ASI-Bench~\citep{zhou2026asibench} move toward execution-based or project-level scientific evaluation. However, existing benchmarks still primarily measure whether an agent completes a task, often within a specific domain, curated input setting, predefined environment, or end-to-end success criterion. In contrast, SciWorkBench focuses on workflow-level experimental-science decisions over raw or imperfect data, uncertainty, artifacts, and evidence-supported conclusions across biology, chemistry, materials science, and physics.
\section{A Unified Framework for Scientific Agent Evaluation}
\label{sec:framework}

\begin{figure}[htbp]
    \centering
    \includegraphics[width=\textwidth]{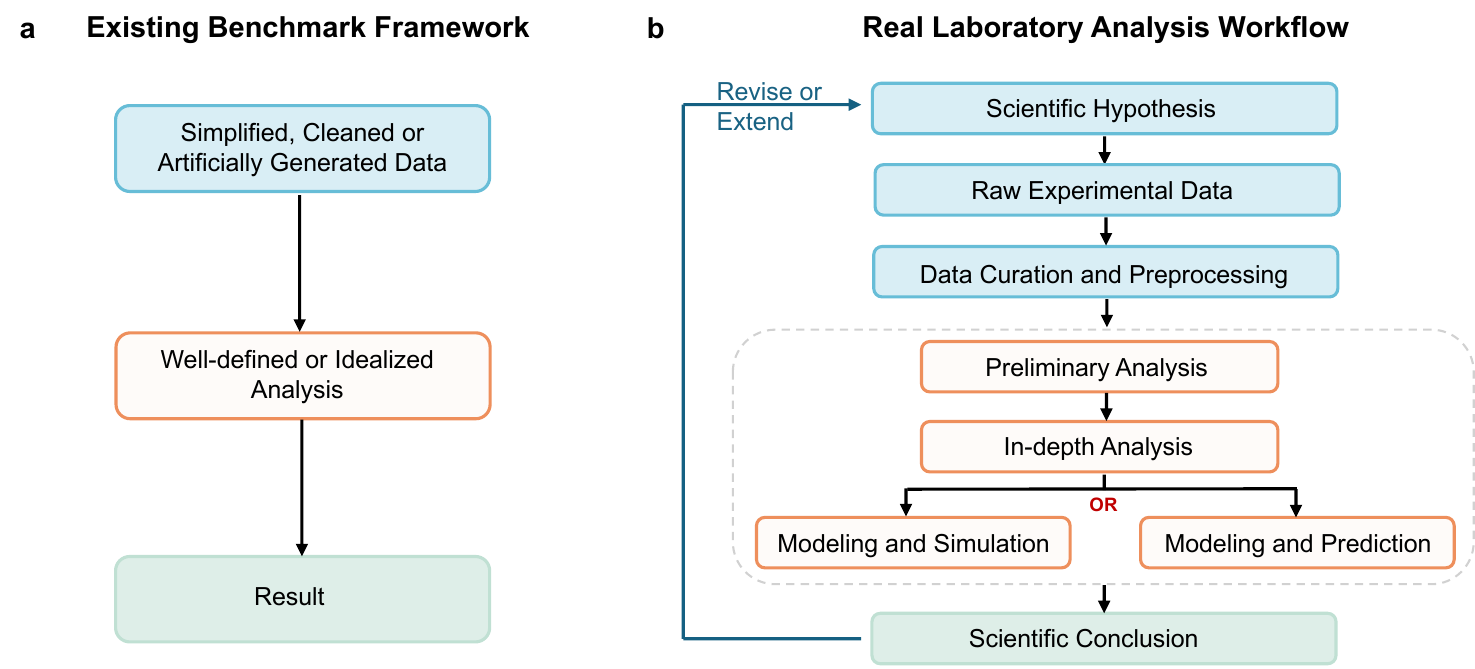}
    \caption{Comparison between existing benchmark framework and our real laboratory analysis framework for scientific-agent evaluation.
    (a) Existing benchmarks often simplify scientific tasks to cleaned inputs, predefined procedures, and final answers.
    (b) Our framework organizes scientific-agent tasks around real experimental workflows.
}
    \label{fig:workflow-framework}
\end{figure}

SciWorkBench is designed to evaluate agents at the level of real laboratory workflows. Existing evaluations often reduce scientific tasks to a simple abstraction: a model receives a cleaned or artificially generated input, executes a well-defined analysis step, and returns an answer (Figure~\ref{fig:workflow-framework}a). However, this simplification removes much of the judgment required in real scientific analysis. The challenge is not only to compute a result, but also to decide what should be analyzed, whether the data are reliable, which assumptions are reasonable, and what conclusions the evidence can support.

We therefore construct a scientific analysis framework around recurring decision points in real experimental analysis (Figure~\ref{fig:workflow-framework}b). Our benchmark is organized around this process. This design better matches how scientists currently use agents in practice: scientists often ask an agent to help solve a specific, realistic, and context-dependent problem within an ongoing research process.

\subsection{Real Laboratory Analysis Workflow}
\label{subsec:workflow-framework}

Our framework abstracts real experimental analysis as an iterative workflow. Tasks in our benchmark are grounded in one or several concrete stages of this workflow, such as inspecting raw experimental outputs, cleaning and preprocessing data, conducting preliminary exploration or in-depth analysis, running simulations or predictions, or interpreting whether the available evidence supports a scientific conclusion. 

Under this setting, the expected output of a task is not merely a numerical answer or an executable script. Depending on the workflow stage being evaluated, the agent must provide an answer in an appropriate scientific form and justify it with the evidence, assumptions, and reasoning required for the conclusion to be scientifically defensible.

This workflow was validated by domain experts. Although these stages are shared across disciplines, their analysis methods and strategies vary substantially by field. For example, data curation in cell biology may involve removing cell-level instrument noise; in diffraction or microscopy, it may involve background correction; and in physical measurements, it may require checking units and instrument calibration. By separating shared workflow stages from domain-specific implementations, this framework supports both cross-domain comparison and fine-grained diagnosis of agent failures.

\subsection{Comparison with Existing Benchmarks}
\label{subsec:benchmark-comparison}

Table~\ref{tab:benchmark-comparison} summarizes why existing benchmarks are not enough to evaluate agents as realistic collaborators in experimental science. Most existing benchmarks replace real scientific workflows with artificial or highly structured proxy tasks. Some reduce science to expert questions, code generation, or result reproduction. Others rely on science-themed puzzles, terminal operations, predefined tools, or curated inputs. These settings may test specific abilities, but they remove the messy data, uncertain analysis choices, and evidence-based judgments that define real experimental research.

\begin{table}[t]
    \centering
    \caption{Comparison between recent scientific agent benchmarks.}
    \label{tab:benchmark-comparison}
    \small
    \hypersetup{hidelinks}
    \setlength{\tabcolsep}{5pt}
    \renewcommand{\arraystretch}{1.2}
    \resizebox{\textwidth}{!}{%
    \begin{tabular}{@{}lccccc@{}}
        \toprule
        \multirow{3}{*}[-2pt]{\textbf{Benchmark}} &
        \multicolumn{2}{c}{\textbf{Evidence source}} &
        \multicolumn{2}{c}{\textbf{Task authenticity}} &
        \textbf{Reasoning} \\
        \cmidrule(lr){2-3}\cmidrule(lr){4-5}\cmidrule(l){6-6}
        & \textbf{Peer-reviewed} & \textbf{Raw} & \textbf{Real experimental}
            & \textbf{Excludes proxy} & \textbf{Evidence-based} \\
        & \textbf{source} & \textbf{observations} & \textbf{workflow}
            & \textbf{puzzles} & \textbf{decisions} \\
        \midrule
        \begin{tabular}[c]{@{}l@{}}LAB-Bench\\[-1pt]{\scriptsize \citep{laurent2024labbench}}\end{tabular}
            & $\triangle$ & $\times$ & $\times$ & $\triangle$ & $\triangle$ \\
        \addlinespace[2pt]
        \begin{tabular}[c]{@{}l@{}}CORE-Bench\\[-1pt]{\scriptsize \citep{siegel2024corebench}}\end{tabular}
            & $\checkmark$ & $\triangle$ & $\triangle$ & $\checkmark$ & $\triangle$ \\
        \addlinespace[2pt]
        \begin{tabular}[c]{@{}l@{}}GeneBench-Pro\\[-1pt]{\scriptsize \citep{li2026genebenchpro}}\end{tabular}
            & $\triangle$ & $\triangle$ & $\triangle$ & $\checkmark$ & $\triangle$ \\
        \addlinespace[2pt]
        \begin{tabular}[c]{@{}l@{}}SciAgentGym\\[-1pt]{\scriptsize \citep{shen2026sciagentgym}}\end{tabular}
            & $\triangle$ & $\times$ & $\triangle$ & $\triangle$ & $\triangle$ \\
        \addlinespace[2pt]
        \begin{tabular}[c]{@{}l@{}}EdgeBench\\[-1pt]{\scriptsize \citep{zhu2026edgebench}}\end{tabular}
            & $\times$ & $\times$ & $\times$ & $\times$ & $\times$ \\
        \addlinespace[2pt]
        \begin{tabular}[c]{@{}l@{}}Terminal-Bench-Science\\[-1pt]{\scriptsize \citep{tbenchscience2026}}\end{tabular}
            & $\times$ & $\triangle$ & $\triangle$ & $\times$ & $\times$ \\
        \midrule
        \begin{tabular}[c]{@{}l@{}}\textbf{SciWorkBench}\\[-1pt]{\scriptsize (Ours, 2026)}\end{tabular}
            & $\checkmark$ & $\checkmark$ & $\checkmark$ & $\checkmark$ & $\checkmark$ \\
        \bottomrule
    \end{tabular}%
    }
    \vspace{4pt}

    \begin{minipage}{\textwidth}
        \fontsize{7.5}{9}\selectfont
        \raggedright
        \emph{Notation.} $\checkmark$, $\triangle$, and $\times$ denote explicit, partial, and little or no coverage, respectively.
        \par\vspace{2pt}
        \emph{Excludes proxy puzzles} indicates whether a benchmark avoids mathematical, programming, or logic-style proxy tasks merely framed in a scientific context.
        \emph{Evidence-based decisions} require agents to reason under noise, artifacts, uncertainty, or competing explanations.
    \end{minipage}
\end{table}

In contrast, SciWorkBench is built from peer-reviewed experimental data and real laboratory scenarios. It exposes agents to raw or imperfect experimental inputs and evaluates them at the level of the experimental workflow. Agents must decide how to process the data, which evidence is reliable, when uncertainty or artifacts limit interpretation, and what conclusions the evidence can support. This makes SciWorkBench both more realistic and more diagnostic: it measures end-to-end performance while also revealing failures that are hidden by evaluations based only on final answers, code execution, tool-use success, or terminal-task completion.
\section{SciWorkBench: Construction and Composition}
\label{sec:benchmark}

\subsection{Task Collection and Quality Control}
\label{subsec:task-collection}

\begin{figure}[htbp]
    \centering
    \includegraphics[width=\linewidth]{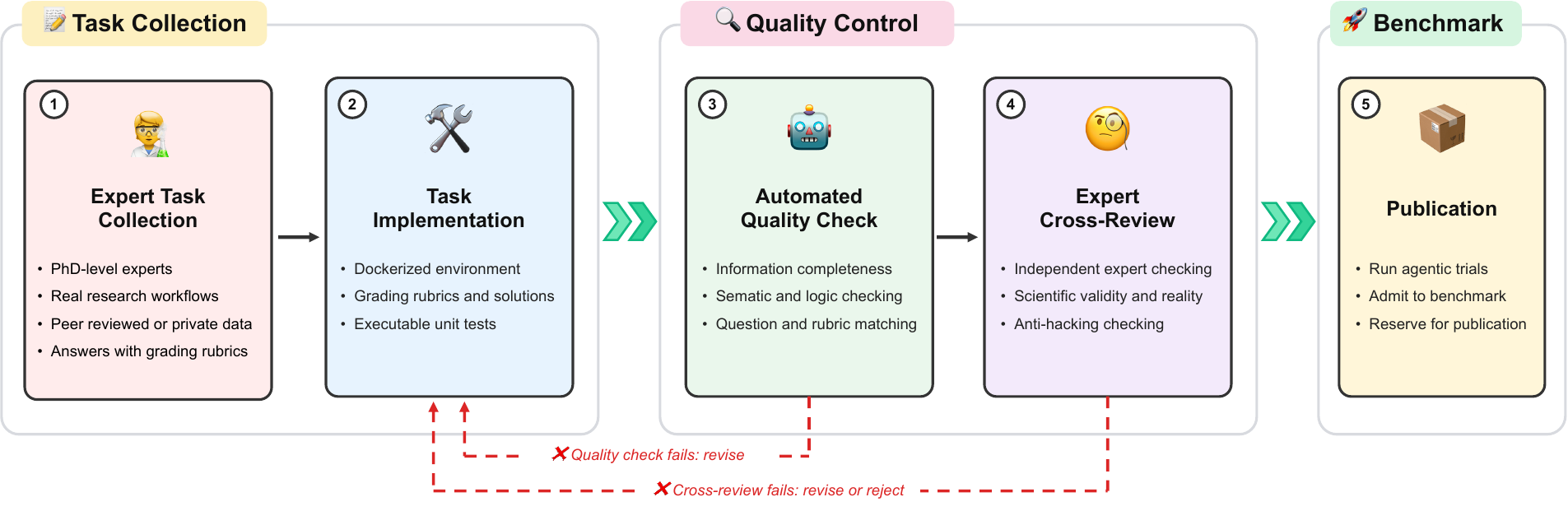}
    \caption{Construction pipeline of SciWorkBench. Tasks are collected from real laboratory scenario, then implemented under dockerized environments, graded with human-provided rubrics, double-checked by both automated workflow and human expert before accepted and released.
    }
    \label{fig:pipeline}
\end{figure}

SciWorkBench is constructed as a set of executable scientific-workflow tasks grounded in real scientific experiments. As shown in Figure~\ref{fig:pipeline}, we follow an expert-driven construction pipeline. Domain experts first identify representative tasks from real research workflows, using data from peer-reviewed publications or expert-contributed private research projects. Each task follows a unified format, consisting of a task prompt, input files, reference answers, and grading rubrics.

To ensure task quality, we apply both automated and expert-based quality control. Automated checks examine information completeness, consistency between the task prompt and grading rubric, and basic semantic or logical errors. Each task is then cross-reviewed by independent experts. The review checks scientific validity, realism, specification completeness, and robustness against shortcut solutions. Tasks that fail these checks are revised or rejected before inclusion. Finally, we run pilot evaluations with agents to ensure that each task can be completed and graded within our evaluation setup.

Across the full construction pipeline, we received 647 candidate tasks, of which only 98 (15.1\%) passed all quality-control stages and were ultimately admitted to SciWorkBench. This retention rate reflects our stringent inclusion criteria for scientific validity, realism, completeness, and reliable execution and grading.

\subsection{Task Format and Example}
\label{subsec:task-format}

\begin{figure}[htbp]
    \centering
    \includegraphics[width=\linewidth]{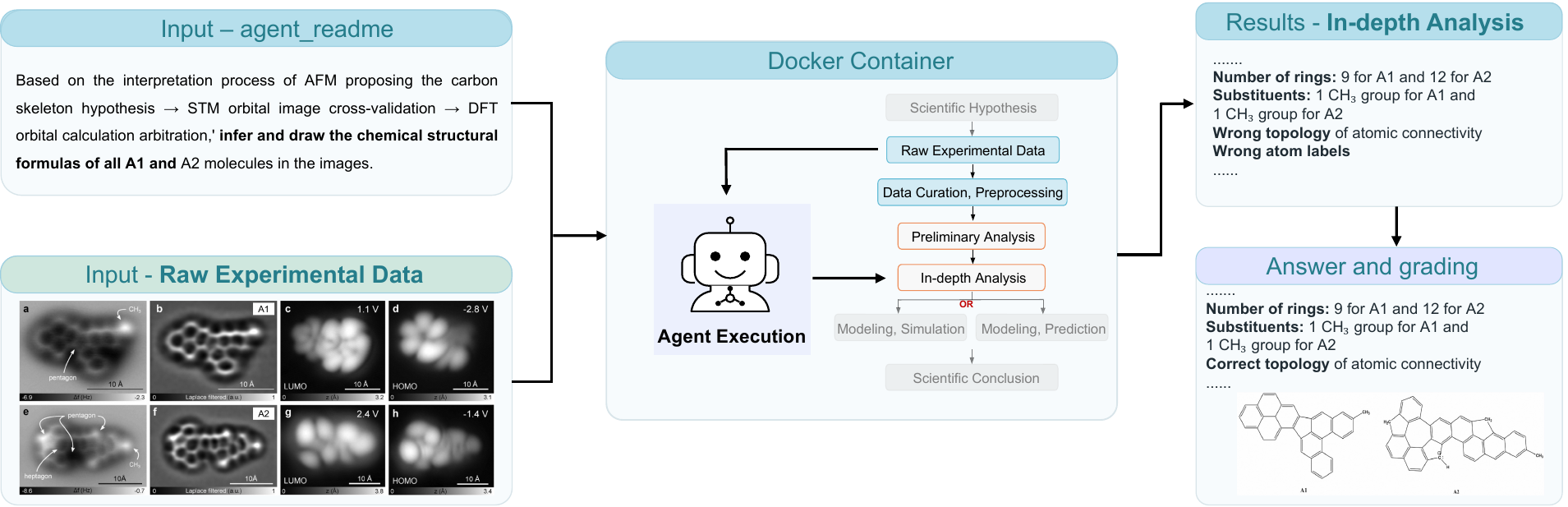}
    \caption{An example from SciWorkBench. Every task in SciWorkBench may cover one or more stages of the proposed experimental-workflow framework.}
    \label{fig:task-example}
\end{figure}

\begin{figure}[htbp]
    \centering
    \includegraphics[width=0.65\linewidth]{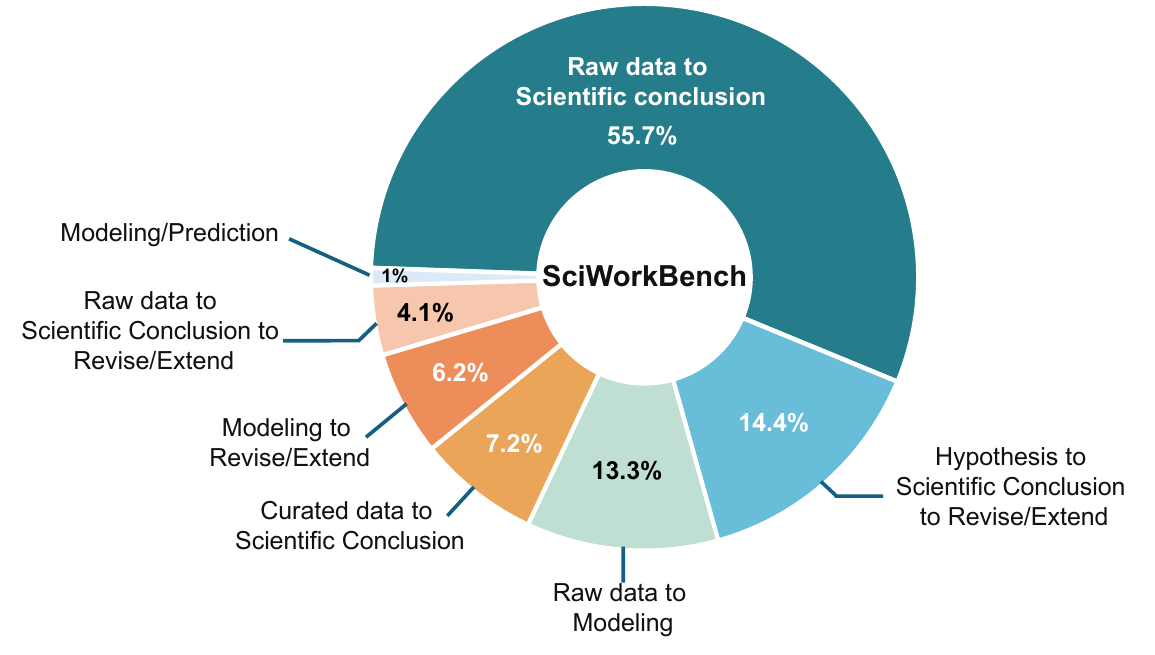}
    \caption{
    Distribution of SciWorkBench tasks by covered workflow stages.
    }
    \label{fig:task_distribution_workflow}
\end{figure}

Each SciWorkBench task provides an agent with the information that a scientist would normally use to solve a concrete research problem. The task prompt describes the scientific context, but does not prescribe a fixed analysis path. The agent must decide how to inspect the data, choose appropriate computational steps, check intermediate results, and produce an answer that is both verifiable and scientifically justified. Figure~\ref{fig:task-example} shows a representative task. A task in SciWorkBench may correspond to one stage or several stages of the workflow framework introduced in Section~\ref{sec:framework}, as shown in Fig.~\ref{fig:task_distribution_workflow}. This design reflects how scientists currently use agents in practice: they typically ask an agent to help with a contextualized problem within an ongoing research process.

\subsection{Disciplinary Distribution of SciWorkBench}
\label{subsec:benchmark-composition}

\begin{figure}
    \centering
    \includegraphics[width=\linewidth]{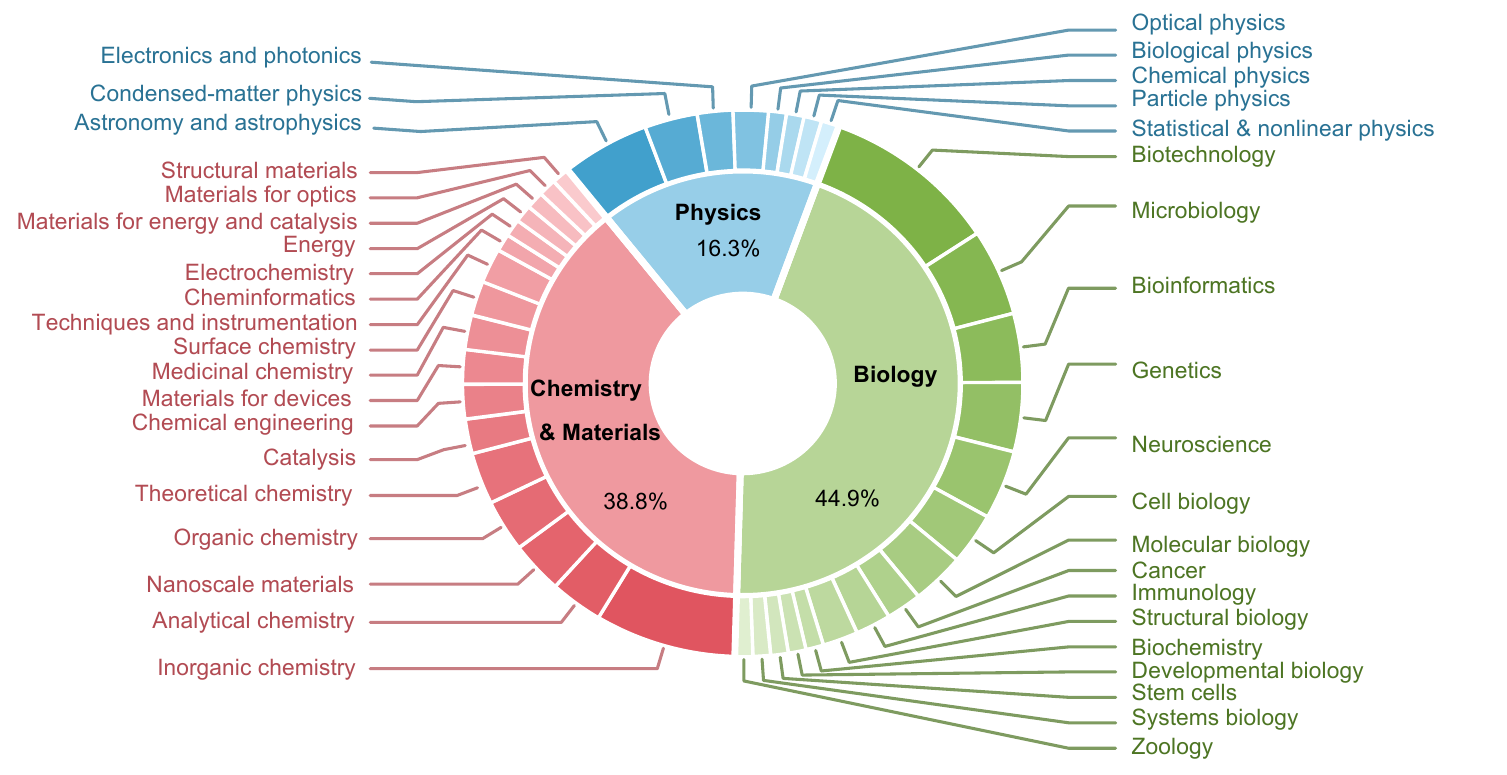}
    \caption{
    Domain and subdomain distribution of tasks in SciWorkBench.
    }
    \label{fig:discipline}
\end{figure}

SciWorkBench covers four major experimental-science domains: biology, chemistry, materials science, and physics. As shown in Figure~\ref{fig:discipline}, the benchmark spans a broad set of subdisciplines, including molecular biology, cell biology, microbiology, neuroscience, immunology, organic chemistry, analytical chemistry, catalysis, electrochemistry, nanoscale materials, structural materials, condensed-matter physics, optical physics, biological physics, and astronomy and astrophysics. 
This distribution is designed to provide broad coverage of common experimental-science subfields across the four domains. It enables both aggregate evaluation and fine-grained diagnosis: aggregate scores measure whether agents can generalize across experimental domains, while subdiscipline-level analyses reveal where agents fail when facing different forms of scientific evidence.

\section{Experiments and results}
\label{sec:experiments}

\subsection{Evaluation Setup}
\label{subsec:setup}

We evaluate 13 agents on all 98 SciWorkBench tasks. Each agent is a pairing of an agent harness with a model. The two harnesses are a Claude Code harness and a Codex harness; the evaluated models include Opus 4.8, Opus 5, GPT-5.5, GPT-5.6 Sol, Qwen3.8-Max, GLM-5.3-Flash, Kimi K3, Seed2.1 Pro, and Gemini 3.5 Flash. All models are evaluated using thinking-depth settings listed in Table~\ref{tab:default-thinking-depth}. Agents receive real scientific task packages from SciWorkBench and execute in a dockerized environment without access to reference solutions. The dockerized environment is preconfigured with all libraries required for the task. Agents are not allowed to access the Internet. Their outputs are stored in the \texttt{output/} directory for later scoring and analysis. Examples of the inputs given to agents are provided in the Appendix.

\subsection{Scoring}
\label{subsec:scoring}

We score task performance using a rubric-based evaluator implemented as Codex with GPT-5.6 Sol. For each task, the evaluator applies rubrics to the agent outputs, rather than task-specific scripts or exact matching. This design reflects the characteristics of scientific workflow tasks: science discovery process often requires interpretation, justified assumptions, and domain-appropriate reasoning, and therefore cannot always be reduced to a single numeric answer. Compared with conventional benchmarks based on fixed scoring rules, our rubric-based assessment provides a more faithful measure of whether an agent's output is well supported and scientifically appropriate.

To reduce potential bias, the judging process is anonymized. The trajectory files do not contain model-identifying information; therefore the evaluator does not know which model is being judged. The testing and judging environments are also kept completely separate, so the evaluator judges only the submitted workflow and final outputs, without access to the original execution environment or model metadata.

Table~\ref{tab:scoring-protocol-comparison} compares different scoring protocols for scientific tasks. Exact-match and script-based scoring are effective when tasks have deterministic outputs, but are poorly suited to tasks involving open-ended analysis, scientific judgment, or multiple valid solution paths. In contrast, agent evaluation using 
expert-authored rubrics better captures whether the outputs satisfy the scientific intent of the task.

\begin{table}[t]
\centering
\small
\caption{Compatibilities of scoring protocols for different science tasks.}
\label{tab:scoring-protocol-comparison}
\setlength{\tabcolsep}{4pt}
\begin{tabular}{lcccc}
\toprule
\textbf{Task type}
& \textbf{Exact match}
& \textbf{Script}
& \textbf{Agent judge} \\
\midrule
Information retrieval tasks
& High
& High
& Medium \\

Structured computational science tasks
& Low
& High
& Medium \\

Scientific workflow tasks
& Low
& Medium
& High \\
\bottomrule
\end{tabular}

\end{table}

\subsection{Overall agent performance}
\label{subsec:results}

\begin{figure}
\centering
\includegraphics[width=\linewidth]{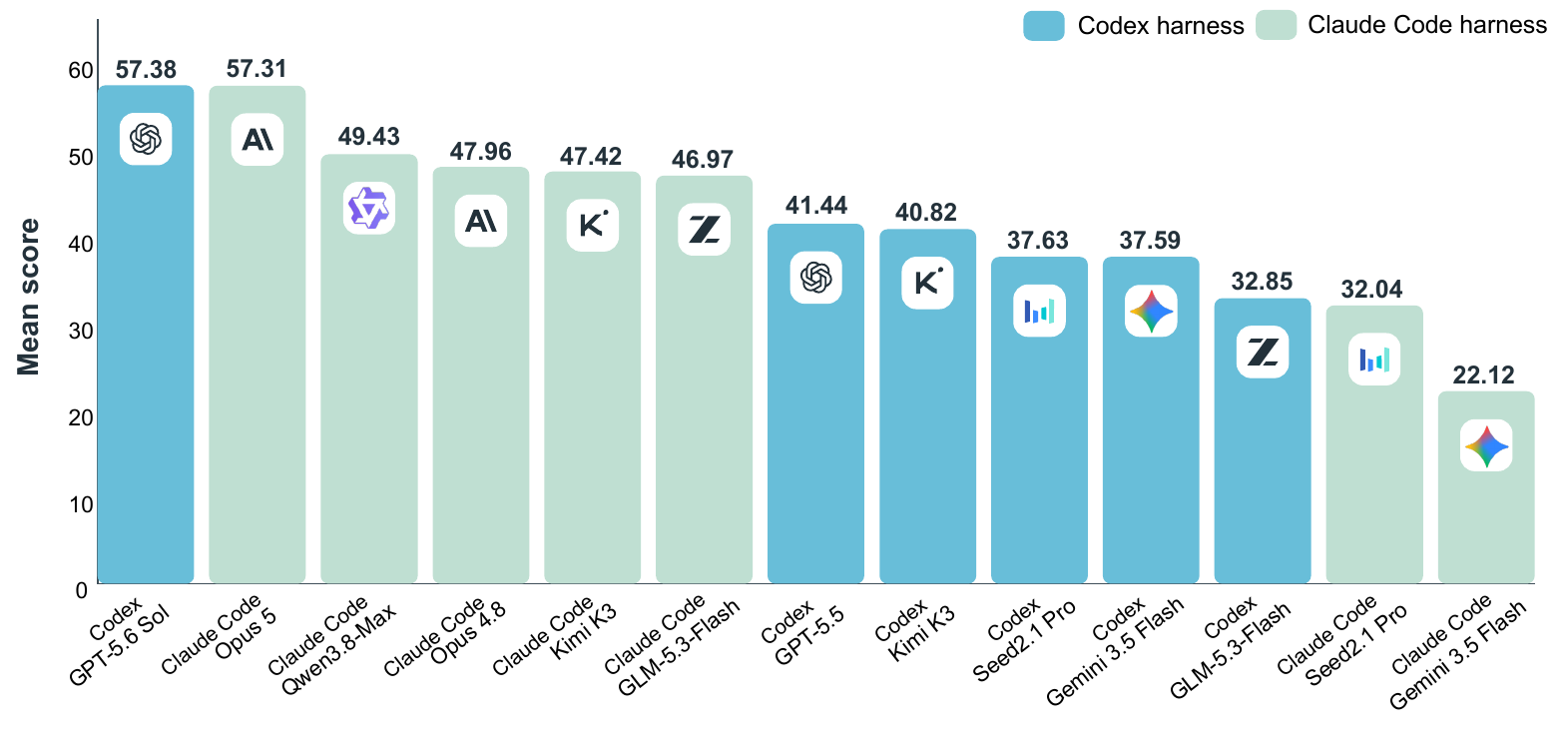}
\caption{
\textbf{Agent performances on SciWorkBench.}
Each column represents one model--harness pair. Scores are mean score on 98 agent tasks from SciWorkBench. Blue indicates Codex harness while light green indicates Claude Code harness.
}
\label{fig:model-harness-ranking}
\end{figure}

Using the evaluation setup and scoring protocol described above, we evaluate the performance of all agents on the 98 SciWorkBench tasks. Here, each agent corresponds to a specific model--harness configuration rather than a base model. As shown in Figure~\ref{fig:model-harness-ranking}, the best-performing agent is Codex with GPT-5.6 Sol, achieving a mean score of 57.38, closely followed by Claude Code with Opus 5 at 57.31. The third-best agent is Claude Code with Qwen3.8-Max, which achieves a mean score of 49.43. Across all evaluated agents, mean scores range from 22.12 to 57.38, indicating that SciWorkBench remains challenging even for the strongest agents. The large spread across model--harness configurations further suggests that scientific workflow performance depends not only on the underlying model, but also on the harness used to run and coordinate the agent.

\subsection{Harness Effect Is Model-Dependent}
\label{subsec:harness-effect}

To further isolate the effect of the agent harness, we compare the four models that were evaluated under both harnesses. We define $\Delta$ as the score under Claude Code minus the score under Codex. As shown in Table~\ref{tab:harness-paired}, the harness effect is strongly model-dependent: the same harness does not consistently improve or degrade performance across models. These results indicate that the model's score is not meaningful unless its harness is specified. Harness and model therefore cannot be treated as independent factors.

\begin{table}[t]
\centering
\small
\caption{Comparison of model performances under different harnesses}.
\label{tab:harness-paired}
\begin{tabular}{lccc}
\toprule
\textbf{Model} & \textbf{Claude Code} & \textbf{Codex} & \textbf{$\Delta$} \\
\midrule
GLM-5.3-Flash       & 46.97 & 32.85 & $+14.12$ \\
Kimi K3             & 47.42 & 40.82 & $+6.60$ \\
Gemini 3.5 Flash    & 22.12 & 37.59 & $-15.47$ \\
Seed2.1 Pro         & 32.04 & 37.63 & $-5.59$ \\
\bottomrule
\end{tabular}
\end{table}

\subsection{Discipline-specific Performance}
\label{subsec:discipline}

We also analyze performance by scientific discipline. Table~\ref{tab:discipline-agent-performance} shows that the leading agent changes across fields. Codex with GPT-5.6 Sol performs best in biology, with a mean score of 54.46, and remains competitive in chemistry/materials and physics. Claude Code with Opus 5 leads chemistry and materials science, scoring 67.59. Claude Code with Qwen3.8-Max leads physics, scoring 57.23.

Across disciplines, the harness effect remains model-dependent. Codex has a slight average advantage in biology, while Claude Code performs better on average in chemistry/materials and physics (Appendix Table~\ref{tab:discipline-harness-average}). The paired comparisons in Appendix Tables~\ref{tab:harness-paired-biology}--\ref{tab:harness-paired-physics} show a similar pattern at the model level. GLM-5.3-Flash and Kimi K3 consistently improve under Claude Code, whereas Gemini 3.5 Flash consistently improves under Codex. Seed2.1 Pro changes direction by discipline, favoring Codex in biology and chemistry/materials but Claude Code in physics. These results show that harness effects depend on both the model and the scientific domain.

\begin{table}[t]
\centering
\small
\caption{Domain-level performance by agent. The best score in each domain is bolded.}
\label{tab:discipline-agent-performance}
\setlength{\tabcolsep}{5pt}
\begin{tabular}{lccc}
\toprule
\textbf{Agent} & \textbf{Biology} & \textbf{Chemistry/Materials} & \textbf{Physics} \\
\midrule
Codex / GPT-5.6 Sol              & \textbf{54.46} & 61.18 & 56.17 \\
Claude Code / Opus 5             & 51.52 & \textbf{67.59} & 51.73 \\
Claude Code / Qwen3.8-Max        & 44.38 & 51.99 & \textbf{57.23} \\
Claude Code / Opus 4.8           & 43.10 & 55.78 & 42.76 \\
Claude Code / Kimi K3            & 45.71 & 51.69 & 42.01 \\
Claude Code / GLM-5.3-Flash      & 43.47 & 52.75 & 43.39 \\
Codex / GPT-5.5                  & 43.15 & 40.82 & 38.21 \\
Codex / Kimi K3                  & 41.15 & 42.20 & 36.61 \\
Codex / Seed2.1 Pro              & 39.52 & 40.54 & 25.50 \\
Codex / Gemini 3.5 Flash         & 36.50 & 39.79 & 35.57 \\
Codex / GLM-5.3-Flash            & 34.31 & 32.99 & 28.51 \\
Claude Code / Seed2.1 Pro        & 34.40 & 29.97 & 30.49 \\
Claude Code / Gemini 3.5 Flash   & 21.22 & 24.77 & 17.50 \\
\bottomrule
\end{tabular}
\end{table}

\subsection{Failure Modes}
\label{subsec:failures}

We analyze the failure cases observed during evaluation and summarize three failure patterns. Failure examples and detailed analyses of these failure causes are provided in Appendix~\ref{app:additional-failure-analysis}.

\paragraph{Mismatch between knowledge and procedure.}
Agents often recognize relevant scientific knowledge, but do not reliably transfer them into task-specific procedures. In some cases, they apply a plausible but incorrect workflow because they fail to align the method with the structure of the data. These failures suggest that scientific knowledge alone is insufficient unless it can be organized into the correct analyzing workflow.

\paragraph{Weak cross-modal evidence integration.}
Many scientific tasks require evidence from text, numerical data, plots, images, and structural diagrams to be used together. Agents can only process one modality at a time, but struggle to assign different modalities the right roles at different stages of the workflow. When agents rely too heavily on a single modality, or use the right modality at the wrong stage, perceptual errors can propagate into incorrect quantification and interpretation.

\paragraph{Limited self-correction.}
Agents also show limited ability to revise earlier decisions. Their self-checking often focuses on whether code runs, files are produced, or formatting requirements are satisfied, rather than whether the result is consistent with domain available evidence. As a result, agents may continue along an invalid analysis path instead of backtracking to reconsider an upstream choice. This indicates that agents for science discovery need self-correction mechanisms based on scientific plausibility and evidential consistency.





\subsection{Limitations}
\label{subsec:limitations}

Although SciWorkBench spans multiple scientific domains and workflow stages, it cannot fully cover the full breadth of science. In addition, due to the high cost of evaluating long-horizon scientific agents, each task--agent setting has limited repetition, making it difficult to fully eliminate variability in agent scores across runs. Finally, rubric-based grading is necessary for open-ended scientific workflows, but future work could further strengthen it through additional grader validation and more detailed checks.

\section{Conclusion}
\label{sec:conclusion}

In this work, we first propose a unified framework for scientific agent evaluation. The framework organizes experimental scientific analysis into a workflow, including hypothesis formulation, raw data inspection, data curation and preprocessing, preliminary analysis, further analysis, modeling or prediction, scientific conclusion, and iterative revision. This framework is intended to capture the workflow-level decisions that scientists make when moving from experimental evidence to defensible conclusions.

Based on this framework, we construct SciWorkBench, a benchmark for evaluating frontier agents on real experimental-science tasks across biology, chemistry, materials science, and physics. Unlike existing benchmarks that focus mainly on isolated knowledge retrieval or single-step computational tasks, SciWorkBench evaluates whether agents can tackle real science questions following scientific workflows.

Our evaluation of 13 model--harness agents shows that they remain far from being able to complete end-to-end scientific agentic tasks reliably. Agent performance varies between models, harnesses, and disciplines, suggesting that science capacity of agents should be evaluated as a system-level property. Our failure analysis also demonstrates agent workflow-level weaknesses, including mismatch between scientific knowledge and task-specific procedures, weak cross-modal evidence integration, and limited scientific self-correction.

SciWorkBench shows that today’s AI agents may seem broadly capable, but they still struggle to reliably finish real experimental workflows. Current limitations do not stem from a single missing capability, but rather from uneven performance across planning, scientific reasoning, tool use, execution, and final delivery. More broadly, SciWorkBench also suggests that improving agent capability requires progress not only in the models, but also in the systems around them. Better models alone will not be enough to make these agents dependable scientific collaborators. They also need better scaffolds to use tools reliably, keep track of intermediate evidence, meet task requirements, and know when the available data are not sufficient to support a conclusion. In this sense, the main challenge is not simply to make agents produce scientifically plausible answers, but to make the full process from raw data to a defensible and reproducible conclusion reliable.

Overall, SciWorkBench highlights both the promise and the current limitations of LLM agents in scientific research. Further progress will require not only stronger models, but also agent systems that can translate scientific knowledge into valid procedures, integrate evidence across modalities, preserve and validate intermediate artifacts, and revise their decisions when results conflict with scientific constraints. The proposed framework and SciWorkBench will provide a practical foundation for evaluating and developing more reliable AI systems for scientific discovery.

\section*{Data and Code Availability}

The benchmark execution and evaluation code is not publicly released. The supplementary material in Appendix~\ref{app:representative-full-task-package} documents an example task, including its instructions, input-data organization, reference-answer structure, and grading rubric. The manuscript source archive includes the instruction and rubric files used to typeset this example. It does not include the complete benchmark data, evaluation environments, agent trajectories, or evaluation and scoring implementation. The example documents the task format and is not a standalone executable benchmark release. Availability of third-party experimental data is subject to the licenses and permissions of the original sources.

\clearpage
\bibliographystyle{plainnat}
\bibliography{references}

\appendix
\clearpage

\setcounter{table}{0}
\setcounter{figure}{0}

\renewcommand{\thetable}{A.\arabic{table}}
\renewcommand{\thefigure}{A.\arabic{figure}}

\renewcommand{\theHtable}{appendix.\arabic{table}}
\renewcommand{\theHfigure}{appendix.\arabic{figure}}

\section{Benchmark Coverage}
\label{app:additional-results}

\subsection{Experimental and Analytical Methods Covered by SciWorkBench}
\label{app:method-word-cloud}

\begin{figure}[!htbp]
    \centering
    \includegraphics[width=\linewidth]{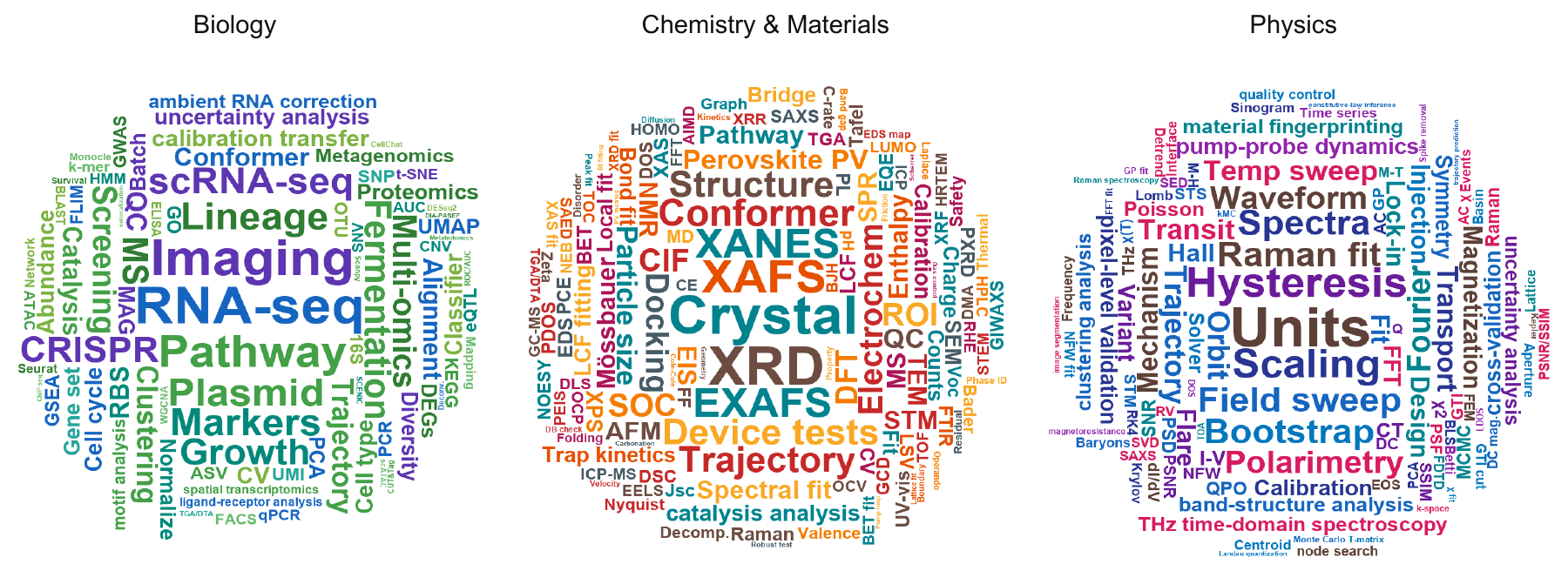}
    \caption{
    Distribution of experimental techniques and analytical methods covered by SciWorkBench tasks.
    }
    \label{fig:appendix-word-cloud}
\end{figure}

To provide an intuitive overview of the scientific tasks included in SciWorkBench, we visualize the experimental techniques and analytical methods mentioned across the benchmark tasks in Figure~\ref{fig:appendix-word-cloud}. The word cloud highlights the diversity of methods covered by the benchmark, ranging from experimental characterization and data processing to modeling, simulation, and quantitative analysis.

\FloatBarrier

\section{Agent Configurations}
\label{app:agent-configurations}

\begin{table}[!htbp]
\centering
\small
\caption{Thinking depth of agent models.}
\label{tab:default-thinking-depth}
\begin{tabular}{ll}
\toprule
\textbf{Model} & \textbf{Thinking depth} \\
\midrule
Opus 4.8 & high \\
Opus 5 & high \\
GPT-5.5 & medium \\
GPT-5.6 Sol & max \\
Qwen3.8-Max & xhigh \\
GLM-5.3-Flash & max \\
Kimi K3 & max \\
Seed2.1 Pro & high \\
Gemini 3.5 Flash & medium \\
\bottomrule
\end{tabular}
\end{table}
\FloatBarrier

\clearpage
\section{Additional Results}
\label{app:discipline-harness-comparisons}

\begin{table}[!htbp]
\centering
\small
\caption{Average discipline-level performance under different harnesses. The best score in each discipline is bolded.}
\label{tab:discipline-harness-average}
\begin{tabular}{lccc}
\toprule
\textbf{Harness} & \textbf{Biology} & \textbf{Chemistry/Materials} & \textbf{Physics} \\
\midrule
Claude Code & 40.54 & \textbf{47.79} & \textbf{40.73} \\
Codex       & \textbf{41.52} & 42.92 & 36.76 \\
\bottomrule
\end{tabular}
\end{table}

\begin{table}[!htbp]
\centering
\small
\caption{Paired comparison of model performance under different harnesses in biology. Positive $\Delta$ indicates that Claude Code outperforms Codex.}
\label{tab:harness-paired-biology}
\begin{tabular}{lccc}
\toprule
\textbf{Model} & \textbf{Claude Code} & \textbf{Codex} & \textbf{$\Delta$} \\
\midrule
GLM-5.3-Flash    & 43.47 & 34.31 & $+9.16$ \\
Kimi K3          & 45.71 & 41.15 & $+4.56$ \\
Gemini 3.5 Flash & 21.22 & 36.50 & $-15.28$ \\
Seed2.1 Pro      & 34.40 & 39.52 & $-5.12$ \\
\bottomrule
\end{tabular}
\end{table}

\begin{table}[!htbp]
\centering
\small
\caption{Paired comparison of model performance under different harnesses in chemistry and materials. Positive $\Delta$ indicates that Claude Code outperforms Codex.}
\label{tab:harness-paired-chemistry-materials}
\begin{tabular}{lccc}
\toprule
\textbf{Model} & \textbf{Claude Code} & \textbf{Codex} & \textbf{$\Delta$} \\
\midrule
GLM-5.3-Flash    & 52.75 & 32.99 & $+19.76$ \\
Kimi K3          & 51.69 & 42.20 & $+9.49$ \\
Gemini 3.5 Flash & 24.77 & 39.79 & $-15.02$ \\
Seed2.1 Pro      & 29.97 & 40.54 & $-10.57$ \\
\bottomrule
\end{tabular}
\end{table}

\begin{table}[!htbp]
\centering
\small
\caption{Paired comparison of model performance under different harnesses in physics. Positive $\Delta$ indicates that Claude Code outperforms Codex.}
\label{tab:harness-paired-physics}
\begin{tabular}{lccc}
\toprule
\textbf{Model} & \textbf{Claude Code} & \textbf{Codex} & \textbf{$\Delta$} \\
\midrule
GLM-5.3-Flash    & 43.39 & 28.51 & $+14.88$ \\
Kimi K3          & 42.01 & 36.61 & $+5.40$ \\
Gemini 3.5 Flash & 17.50 & 35.57 & $-18.07$ \\
Seed2.1 Pro      & 30.49 & 25.50 & $+4.99$ \\
\bottomrule
\end{tabular}
\end{table}
\FloatBarrier

\clearpage
\section{Example Task}
\label{app:representative-full-task-package}

We present one example task to illustrate the concrete structure of SciWorkBench tasks. During evaluation, the agent receives the instruction file and the data directory. The reference answer package and grading rubrics are visible only to the agent responsible for grading and are included here only to make the task format transparent.

\subsection{Instructions: \texttt{agent\_readme.md}}

\begin{Verbatim}[
fontsize=\scriptsize,
breaklines=true,
breaksymbolleft={},
breaksymbolright={},
frame=single,
framesep=2mm
]
# Agent Task: Selecting a Stability-Control Model for Acidic OER on RuO₂

## 1. Scientific background

Ruthenium oxides are highly active catalysts for the acidic oxygen evolution reaction, but their stability can vary substantially with hydration state, structural order, and electrode configuration. This task provides a complete, reorganized dataset from a public article. The data span electrochemical performance, long-term operation, elemental dissolution, spectroscopy, thermal desorption, mass-spectrometric depth analysis, diffraction, surface properties, and atomic coordinates.

Your job is not to repeat the article’s conclusion. You must independently analyze the complete dataset and determine which candidate model explains the observed stability differences with the greatest coverage and the fewest contradictions.

## 2. Input data

All inputs are under `data/`. They are derived from DOI `10.1038/s41467-025-56188-z` and are not personal experimental data.

The package contains:

- complete CSV transcriptions of all 26 Source Data worksheets;
- a long-form table of every nonempty worksheet cell;
- 72 paired numerical curves and a series catalog;
- a complete catalog of Supplementary Figures S1–S24 and Tables S1–S4, together with every publicly available numerical transcription;
- two complete atomic-coordinate structures;
- provenance, schema, coverage, and checksum records.

Supplementary figures without published raw points provide only factual catalog information or reported values. Do not create artificial numerical precision by digitizing those figures. Every input file must appear in your inventory and receive a relevance assessment, whether or not it ultimately supports your conclusion.

## 3. Single scientific question

Select exactly one of the following mutually exclusive candidate models as the model most consistent with the complete dataset:

| ID | Candidate model |
|---|---|
| M0 | Ru dissolution dominated |
| M1 | Ru-center oxidation-state change dominated |
| M2 | Catalyst-bulk hydration/proton-related process dominated |
| M3 | Catalyst-layer reaction-interface stability dominated |
| M4 | Coupled bulk and reaction-interface processes |

You must return one model ID. If input integrity or data quality prevents a defensible comparison, you may return `INDETERMINATE_DATA_FAILURE`, but you must identify the precise failure.

## 4. Objectives

1. Audit the integrity, readability, provenance, and numerical coverage of every input.
2. Independently select and implement quantitative analyses capable of distinguishing the five candidate models.
3. Keep direct calculations, transcribed reported values, and qualitative evidence clearly separated.
4. For each candidate model, identify supporting evidence, opposing evidence, unresolved issues, and a final disposition.
5. Express how evidence converges on the final model through auditable dependencies rather than a conclusion-only narrative.
6. State the limits of the available evidence and identify quantities that cannot be independently recomputed from the supplied files.

## 5. Deliverables

Place `analysis.py` in the task root. A single run must generate the following files under `output/`:

1. `data_audit.json` — input files, checksums, coverage, and anomalies;
2. `computed_metrics.json` — calculated metrics with units, methods, and source locators;
3. `model_evaluation.csv` — evidence comparison and disposition for M0–M4;
4. `evidence_graph.json` — evidence nodes, dependencies, and the final model;
5. `all_data_inventory.csv` — role and relevance assessment for every file under `data/`;
6. `analysis_summary.png` — an integrated model-comparison figure of at least 1800 × 1200 pixels;
7. `technical_report.md` — the complete technical report.

Minimum machine-readable fields are specified in `data/task_constraints.json`.

## 6. General requirements

- `analysis.py` must run offline, read only `data/`, never read `answer/`, and must not hard-code the final conclusion.
- Every numerical result must state its unit, calculation method, and source file. State any rounding convention.
- `model_evaluation.csv` must contain exactly five rows, one for each of M0–M4.
- Every inference node in `evidence_graph.json` must trace to data, metrics, or earlier nodes.
- `technical_report.md` must begin with the unique conclusion, followed by reproducible evidence, comparison of competing models, and limitations.
- Do not turn correlation into a claim that every alternative microscopic mechanism has been excluded.
- Do not omit data from the inventory or audit merely because it is not used in the final argument.

## 7. Evaluation focus

Evaluation emphasizes data coverage, numerical correctness, fair comparison across models, complete evidence dependencies, reproducibility, and appropriately bounded conclusions. Reference results, tolerances, and scoring logic are not disclosed in the examinee materials.
\end{Verbatim}

\subsection{Input Data Directory: \texttt{data/}}

This task data directory contains a complete reorganized data package derived from the source article. It includes:

\begin{itemize}[leftmargin=*]
    \item complete CSV transcriptions of all 26 source-data worksheets;
    \item a long-form table containing every nonempty worksheet cell;
    \item 72 paired numerical curves and a corresponding series catalog;
    \item supplementary figure and table catalogs, including Supplementary Figures S1--S24 and Tables S1--S4;
    \item transcribed supplementary values for BET, crystallinity, ICP-MS, DFT-reported quantities, literature comparison, reported events, and surface metrics;
    \item two complete atomic-coordinate structures;
    \item provenance, schema, task-constraint, coverage, and checksum files.
\end{itemize}

The data directory is provided to the agent, but the full raw contents are not reproduced in the paper appendix because of length. This summary is included to show the scope and heterogeneity of the supplied evidence.

\subsection{Reference Answer Package: \texttt{answer/}}

The answer directory contains the following files:

\begin{itemize}[leftmargin=*]
    \item {\footnotesize\texttt{reference\_solver.py}}: reference implementation;
    \item {\footnotesize\texttt{reference\_protocol.json}}: reference analysis protocol;
    \item {\footnotesize\texttt{reference\_answer.md}}: reference answer summary;
    \item {\footnotesize\texttt{reference\_output/data\_audit.json}}: reference data audit;
    \item {\footnotesize\texttt{reference\_output/computed\_metrics.json}}: reference computed metrics;
    \item {\footnotesize\texttt{reference\_output/model\_evaluation.csv}}: reference model-comparison table;
    \item {\footnotesize\texttt{reference\_output/evidence\_graph.json}}: reference evidence-dependency graph;
    \item {\footnotesize\texttt{reference\_output/all\_data\_inventory.csv}}: reference full data inventory;
    \item {\footnotesize\texttt{reference\_output/analysis\_summary.png}}: reference summary figure;
    \item {\footnotesize\texttt{reference\_output/technical\_report.md}}: reference technical report.
\end{itemize}

\subsection{Grading Rubrics: \texttt{grading.md}}

\begin{Verbatim}[
fontsize=\scriptsize,
breaklines=true,
breaksymbolleft={},
breaksymbolright={},
frame=single,
framesep=2mm
]
# Result-Only Grading Rubric (100 points)

## 1. Full-data audit (20 points)

- 5 points: all 117 checksum-listed files pass SHA-256 validation.
- 5 points: correctly reports 26 complete worksheets, 72 paired curves, and 232,092 paired curve points.
- 5 points: correctly reports 492,284 source numerical cells from row 3 onward, 35,477 unpaired numerical cells, and zero numerical cells missing from the exhaustive transcription; unpaired cells must not be described as deleted.
- 3 points: correctly reports 24 supplementary figures plus 4 supplementary tables, and two structures totaling 150 atoms.
- 2 points: states that the data are DOI-derived rather than personal experimental data and confirms that publisher attachments are not redistributed as byte copies.

If `data_audit.json` has an overall `pass` value of false but the submission still selects M0–M4, this section receives zero and the final-conclusion score is capped at 2 points.

## 2. Numerical reconstruction (30 points)

- 5 points: reconstructs the nine polarization voltages at the three current densities; all six Ir-minus-Ru voltage differences are positive; the maximum Fig. 1a/Fig. 1b discrepancy is `0.0144881 V` within ±`1e-5 V`.
- 5 points: durations are h-RuOx `64.1 s`, anh-RuO2 `549.0 h`, h-RuOx-300 `133.63333 h`, and anh-RuO2-S `114.9 h`; the converted duration ratio is `30833.07` within 0.1% relative tolerance.
- 4 points: dissolution inventory fraction `0.0005263158`, equivalent to `0.0526316%`.
- 4 points: XANES EoL−BoL shift `0.0677131 eV`; EXAFS Ru–O/Ru–Ru retention ratios `0.926251/0.852349`.
- 5 points: TPD-H2O ratio `64.41668`, ToF-OH ratio `4.083059`, OH-depth mean ratio `1.398199`, and area-specific-capacitance ratio `8.25`, each within 0.5% relative tolerance.
- 4 points: XRD apparent sizes at 300/600/800 °C are `5.04494/18.73697/37.34619 nm`, within 0.5% relative tolerance.
- 3 points: anh-RuO2-S OLS slope `0.495961 mV h^-1` and h300-to-h300-S improvement ratio `10.25`.

## 3. Candidate-model comparison and unique conclusion (25 points)

- 10 points: `model_evaluation.csv` contains exactly five rows, M0–M4; each has supporting evidence, opposing evidence, unresolved issues, and an explicit disposition.
- 8 points: `evidence_graph.json` contains at least 12 substantive nodes; every inference traces to an input, metric, or predecessor, and the dependencies converge on one model.
- 5 points: uniquely selects `M4`, consistently across `model_evaluation.csv`, `evidence_graph.json`, and the report.
- 2 points: M0, M1, M2, and M3 each have a data-driven reason for not being selected; “all are possible” is not accepted.

## 4. Technical report and evidence dependencies (20 points)

- 12 points: at least 12 dependent reasoning nodes cover integrity, performance comparability, time scale, evaluation of M0/M1, independent bulk-associated observables, thermal response, high-current scope, fixation response, and the final comparison.
- 4 points: direct curve calculations, transcribed SI values, qualitative statements, and geometric proxies are explicitly separated.
- 2 points: states that supplementary figures without raw points were not digitized to create precision.
- 2 points: frames the conclusion as the best model within the supplied candidate set and data, not as proof excluding every other microscopic mechanism.

## 5. Inventory, machine-readable outputs, and figure (5 points)

- 2 points: `all_data_inventory.csv` lists all 118 files under `data/`, including the checksum file itself, with a data role and relevance assessment.
- 1 point: every JSON/CSV file satisfies the minimum schema in `task_constraints.json`.
- 2 points: the integrated PNG is at least 1800 × 1200 pixels, covers at least three independent evidence classes, and identifies the selected model clearly.
\end{Verbatim}

\FloatBarrier

\clearpage
\section{Additional Failure Analysis}
\label{app:additional-failure-analysis}

This section provides additional examples and explanations for the failure modes summarized in Section~\ref{subsec:failures}.

\newcolumntype{Y}{>{\raggedright\arraybackslash}X}

Table~\ref{tab:failure-analysis-cases} lists representative failure cases selected from agent execution trajectories. These cases are used as qualitative diagnostic examples rather than estimates of failure-mode frequency.

{\scriptsize
\setlength{\tabcolsep}{3pt}
\renewcommand{\arraystretch}{1.15}
\begin{xltabular}{\textwidth}{@{}>{\raggedright\arraybackslash}p{0.16\textwidth} Y Y >{\raggedright\arraybackslash}p{0.14\textwidth}@{}}
\caption{Representative failure cases across scientific tasks.}
\label{tab:failure-analysis-cases}\\
\toprule
\textbf{Task} & \textbf{Exposed issue in the agent trajectory} & \textbf{Expected workflow} & \textbf{Main failure pattern} \\
\midrule
\endfirsthead

\toprule
\textbf{Task} & \textbf{Exposed issue in the agent trajectory} & \textbf{Expected workflow} & \textbf{Main failure pattern} \\
\midrule
\endhead

gRNA transcriptional valve design &
The agent produced valid DNA constructs, primers, assembly plans, and sequence-level checks. However, the scientific goal was not only to build valid sequences. The next-round design also needed to disentangle valve identity, slot position, upstream RNAP flux, and RNAP dosage. The trajectory therefore exposes a mismatch between low-level construct validation and the higher-level experimental-design objective. &
Integrate array read-depth data, standalone valve measurements, RNAP perturbation data, and assembly constraints to design a small construct set that supports interpretable future comparisons. &
Weak cross-modal evidence integration \\

PBMC doublet / low-quality cell identification &
After dependency and coding failures, the agent relied mainly on heuristic QC and marker-score rules. Its checks confirmed cell coverage, file format, and output completeness, but did not sufficiently test whether the evidence actually separated doublets, degraded cells, and ambiguous biological states. &
Use the count matrix, marker panel, QC metrics, mixed-lineage evidence, and uncertainty labels to distinguish singlet, doublet, low-quality, and uncertain barcodes. &
Limited self-correction \\

diaPASEF metaproteomics &
The task required a real metaproteomics workflow, including spectral-library construction, FDR control, peptide ambiguity handling, quantification, and statistical analysis. After finding that raw \texttt{.d} mass-spectrometry files were unavailable, the agent still constructed a simulated quantitative analysis from FASTA and annotation tables. It recognized relevant scientific concepts but did not respect the procedural boundary of what could be inferred without raw spectra. &
Build spectral libraries, control FDR, remove ambiguous human/microbial peptides, quantify proteins, taxa, and functions, and perform clinical group comparisons. &
Mismatch between knowledge and procedure \\

PCET HEOM simulation &
The agent reported a Fermi--HEOM implementation with MPS/MPO propagation and convergence sweeps over hierarchy depth, bond dimension, and time step. However, several parameter changes produced nearly identical \(P_0(5~\mathrm{fs})\) values. The trajectory suggests that the convergence checks verified that runs completed, but did not verify that the numerical computation was physically sensitive to the intended truncation parameters. &
Implement the specified PCET HEOM model, propagate auxiliary density operators in tensor-network form, and perform verifiable convergence tests over hierarchy, bond dimension, time step, and basis truncation. &
Limited self-correction \\

BaCoO$_3$ TGA/DTA thermochemistry &
The agent performed peak integration, calibration, and Hess-law back-calculation, but the final formation enthalpy depended strongly on assumed oxygen partial pressure, entropy change, baseline choice, and literature calibration anchor. The trajectory emphasized file generation and plot completeness more than checking whether these assumptions dominated the final thermochemical conclusion. &
Identify reaction stages from TGA/DTA, integrate the carbonate peak with justified baselines, calibrate instrument response, apply Kirchhoff and Hess corrections, and quantify assumption sensitivity. &
Limited self-correction \\

Single-crystal X-ray diffraction structure solution &
The task required determining the space group, solving the structure from reflection data, assigning atom types, refining anisotropic displacement parameters, and producing a model that could be recomputed from \texttt{sample.hkl}. The trajectory shows the agent recognized the crystallographic terms and checked tool availability, but the core difficulty was not file generation; it was backtracking from implausible or incomplete structure-solution choices when standard crystallographic software or procedures were unavailable. &
Determine systematic absences and space group, solve the initial structure, refine the model against the official HKL data, validate chemistry and geometry, and export consistent CIF and JSON models. &
Limited self-correction \\

SEM particle-size analysis &
The task depended on visual evidence: interpreting the scale bar, cropping the label region, separating touching particles, and auditing segmentation overlays. The trajectory shows that the agent noticed watershed over-segmentation of elongated particles and adjusted marker generation. This case illustrates how an early segmentation or calibration error can systematically distort downstream particle counts, diameter distributions, IQRs, and particle densities. &
Calibrate the scale bar, remove the annotation region, segment particles using Otsu thresholding, hole filling, and watershed, visually audit overlays, then compute particle-size and density statistics. &
Weak cross-modal evidence integration \\

MPMS3 magnetic hysteresis analysis &
The task required reconstructing raw magnetic measurements, applying background correction, unit conversion, and normalization, and then extracting hysteresis quantities with uncertainty checks. The exposed issue is that agents may recognize the relevance of hysteresis and magnetic correction but still treat the loop as a generic curve-fitting problem, without preserving acquisition order, branch structure, and correction provenance. &
Parse MPMS3 raw records, identify measurement segments, apply background and unit corrections, separate hysteresis branches according to acquisition order, and quantify coercivity, remanence, relaxation, and robustness. &
Mismatch between knowledge and procedure \\

UCXB censored-trajectory prediction &
The agent recognized the overall UCXB prediction task, but made an upstream event-definition error. It should first identify contiguous activity blocks and then merge them using endpoint age gaps. The incorrect block logic split continuous activity segments, contaminating training labels, channel assignments, detachment and recontact times, and downstream probabilistic predictions. &
Merge trajectory shards, remove duplicates, sort by source, model, and age, identify contiguous activity blocks, merge events by the specified rule, then train and predict censored outcomes. &
Mismatch between knowledge and procedure \\

Cryo-EM particle extraction &
The task required CTF estimation, particle picking, particle extraction, and a RELION-compatible \texttt{particles.star}. The trajectory shows missing cryo-EM tools and key Python packages; after installation failed, the agent moved toward a lightweight substitute. The exposed issue is that producing STAR-like files or coordinate outputs can be mistaken for completing the domain workflow, even though CTF estimation and standard particle extraction were not actually performed. &
Read acquisition parameters, estimate CTF for each micrograph, pick particles, extract particle stacks, and validate a RELION 3.1+ two-table STAR file. &
Limited self-correction \\

\bottomrule
\end{xltabular}
}

\paragraph{Failure to use established domain workflows.}
A recurring issue is that agents sometimes replace established scientific procedures with ad hoc implementations. These improvised workflows may appear reasonable at a high level, but they often miss domain-specific constraints that are essential for obtaining valid results. For example, in hysteresis-loop analysis, some agents recognized the relevance of magnetic hysteresis and \(B\)--\(H\) correction, but failed to separate the ascending and descending branches according to the actual acquisition order of the measurements. In other tasks, agents overlooked mature domain tools or validated analysis pipelines and instead constructed fragile custom scripts. Such implementations can introduce subtle errors in preprocessing, parameter handling, normalization, or numerical fitting, which may be sufficient to change the final scientific conclusion.

\paragraph{Limitations in scientific visual interpretation.}
Some failures originate from the agent's limited ability to interpret scientific visual evidence. Scientific figures and images often contain task-critical information that is not explicitly represented in text or tables. We observe cases where agents confuse spectral peaks with noise, fail to trace relevant morphology, or incorrectly detect and segment objects in microscopy images. These perceptual errors are especially consequential because they usually occur early in the workflow and then propagate into downstream measurements, statistical summaries, and scientific interpretation.

\paragraph{Incorrect coordination across modalities.}
Beyond recognizing individual modalities, agents must decide how each modality should be used at different stages of an analysis. This coordination remains difficult. For example, nanoparticle-size analysis may require visual identification of individual particles, pixel-level segmentation based on image contrast, and calibrated measurement of particle dimensions. Some agents instead estimate particle size directly by visual inspection during the quantitative stage, while others rely only on intensity differences and mistakenly include artifacts, background structures, or overlapping supports as particles. These examples suggest that cross-modal scientific reasoning requires not only extracting information from each modality, but also assigning each source of evidence an appropriate role within the workflow.

\paragraph{Insufficient scientific backtracking.}
Agents also tend to continue with their initial analysis path even when intermediate results suggest that an earlier decision may be wrong. In single-crystal X-ray diffraction structure solution and refinement, for instance, a chemically implausible structure or a model that poorly explains the diffraction data should prompt reconsideration of upstream choices, such as systematic-absence analysis, space-group assignment, or the structure-solution strategy. However, agents often treat successful code execution or file generation as sufficient evidence of progress. They may therefore continue refining an invalid model rather than returning to the decision point that caused the inconsistency. This highlights the need for self-correction mechanisms that evaluate intermediate results against scientific constraints, rather than checking only procedural completion.

\end{document}